\documentclass{article}[15pt]

\usepackage{amsmath,epsfig,color,calc,rotating}
\usepackage{graphics}
\usepackage{wrapfig}
\usepackage{ulem}

\usepackage{natbib}
\bibpunct{(}{)}{;}{a}{}{,}

\begin{document}

\title{
Characterizing Ranked Chinese Syllable-to-Character Mapping Spectrum: 
A Bridge Between the Spoken and Written Chinese Language
\vspace{0.2in}
\author{
Wentian Li\\
{\small \sl The Robert S. Boas Center for Genomics and Human Genetics}\\
{\small \sl The Feinstein Institute for Medical Research} \\
{\small \sl North Shore LIJ Health System}\\
{\small \sl Manhasset, 350 Community Drive, NY 11030, USA.}\\
}
\date{}
}
\maketitle  % End title section
\markboth{\sl Li}{\sl Li}

\begin{center}
{\bf ABSTRACT}: 
\end{center}

One important aspect of the relationship between spoken 
and written Chinese is the ranked syllable-to-character
mapping spectrum, which is the ranked list of syllables
by the number of characters that map to the syllable.  
Previously, this spectrum is analyzed for more than 
400 syllables without distinguishing the four intonations.
In the current study, the spectrum with 1280 toned syllables
is analyzed by logarithmic function, Beta rank function,
and piecewise logarithmic function. Out of the three
fitting functions, the two-piece logarithmic function
fits the data the best, both by the smallest sum of
squared errors (SSE) and by the lowest Akaike information
criterion (AIC) value. The Beta rank function is the
close second. By sampling from a Poisson distribution whose 
parameter value is chosen from the observed data, we empirically 
estimate the $p$-value for testing the two-piece-logarithmic-function 
being better than the Beta rank function hypothesis, to be 0.16.
For practical purposes, the piecewise logarithmic function 
and the Beta rank function can be considered a tie.

\large

\section{Introduction}

\indent

Chinese language has been considered to be one of the hardest to learn 
languages for non-natives, or at least, ``strikingly different"
to an European language speaker \citep{wang73}. For spoken 
Chinese, a sound (syllable) consists of a consonant initial 
(e.g. {\sl h}), a vowel final
(e.g. {\sl ao}), and one of the four intonations (e.g.,
the dipping intonation 3: {\sl hao3}).  Although the syllables are now 
written in Roman/Latin alphabets under the {\sl pinyin} system, 
and are pronounced similarly as many Romance/Latin languages,
the four distinct intonations (changing of pitch contour)
do not match a similar system in Western languages.
For written Chinese, the level of degeneracy can be high.
One-to-one correspondence between ideogram or character and syllable is
rare, such as the example of {\sl neng2} which has only one 
character and one meaning of ``can", ``to be able of", 
while tens of characters per syllable is common. 
Besides these complications, each character may have several 
meanings depending on the context.

The many-to-one mapping from a sound in spoken Chinese to
characters in written Chinese crucially contributes to the
complexity of the Chinese language. The term ``polymorphous" can be
used to describe the situation where many characters are mapped
to one syllable. The terms ``homophone" \citep{homop,homop-su} 
and ``heterograph" \citep{heterog-su,heterog-c} used in 
linguistics describe similar situations, though these are used 
at the word level.  Homophome emphasizes the aspect of the same 
pronunciation, whereas heterograph emphasizes the difference in writing.

In order to characterize numerically this polymorphous feature
of Chinese language,  we ask the following question: on 
average, how many written characters correspond to a toned
syllable? The next quantitative linguistic question is: 
if the  number of characters per syllable is ranked 
for all syllables (with intonation), termed ``syllable-to-character 
mapping spectrum", does the decrease of this measurement 
with rank follow a particular functional form?

There are around 400 syllables in spoken Chinese ignoring
intonation (some syllables are rare, and/or used only in a
colloquial form). When intonation is considered, the number 
of syllables should be multiplied by 4, of the order of 1600. 
The number of Chinese characters included in a given dictionary, 
however, is not fixed. Besides a core group of commonly 
used characters, which can be in the range of low thousands 
(e.g. $\sim$ 3000 covered by primary school textbook:
{\sl http://zd.diyifanwen.com/zidian/szb/}), the total 
number can be as high as 50,000 $\sim$ 60,000. This uncertainty
of the total number of Chinese characters is reminiscent of
the uncertainty of the total number of English words described
under the ``large number of rare event" model \citep{baayen}.

Besides these, two more facts should be considered in counting 
the number of characters. First, some characters are only used in classic 
literature while rarely used in modern Chinese. Second, a set of 
simplified characters (roughly 2000) were introduced in 1964 in mainland
China ({\sl http://www.china-language.gov.cn/wenziguifan/managed/002.htm}),
which co-exist with the ``traditional" characters used in Taiwan and
oversea Chinese communities. Also, to consider Chinese as a
single language from both the written and spoken language 
perspective obviously ignores the diversity of spoken Chinese 
dialects. Here we focus only on one spoken and one written Chinese
-- the Manderin (or {\sl putonghua}, or Pekingese) \citep{wang12,shen}.

In a previous study, the ranked syllable-to-character
mapping spectrum is analyzed without considering intonation
\citep{wli-physicaa}. Two dictionaries were used with
very different number of entries: a small dictionary
with 9212 characters, and a larger online dictionary with
21783 characters. Despite the size difference of the
two dictionaries, the ranked syllable-to-character
mapping spectra after normalization are virtually
identical \citep{wli-physicaa}. Based on this observation,
we will only use one dictionary for the current analysis
that includes intonation.

For a ranked syllable-to-character mapping spectrum,
the $x$-axis is the rank of syllables, rank-1 for
the most polymorphous syllable (with the most number
of characters pronounced in the same way), and lowest 
ranks for the less polymorphous syllables. The 
$y$-axis is the normalized number of characters per syllable. 
The main result from \citep{wli-physicaa} is that 
when the $x$-axis is logarithmically transformed, 
the fall-off of the spectrum is close to a straight line, 
indicating a logarithmic functional form. However, the Beta 
rank function \citep{beta1,beta2,beta3} is shown to be 
even better in fitting the data \citep{wli-physicaa}, 
if only slightly better. In this paper, we will examine
whether the same conclusion still holds when
intonation-included syllables are considered.

Different functional forms of syllable-to-character
mapping spectrum provide numeric characterization of
the fall-off from the polymorphic to the monomorphic
(one-to-one) syllables. Such quantitative characterization
should play a similar role as the Zipf's law in
characterizing the word usage in languages \citep{zipf}. 
Not only such function makes the estimation of mean,
median, variance easier, but also it helps to
prioritize the learning of the language, and perhaps
provide hints on the evolution of the interplay between
the written and spoken Chinese.

%\begin{table}
%\begin{center}
%[table 1 here]
%\vspace{0.5in}
%\end{center}
%\caption{
%\label{table1}
%The top-ranking syllable yi4 with 83 characters included
%in the {\sl Modern Chinese Small Dictionary}. These characters
%are further grouped: some are the more common and
%modernly used, some are only used in literature, some
%are for person, place, animal names only, etc.
%}
%\end{table}

\begin{figure}[th]
\begin{center}
  \begin{turn}{-90}
   \epsfig{file=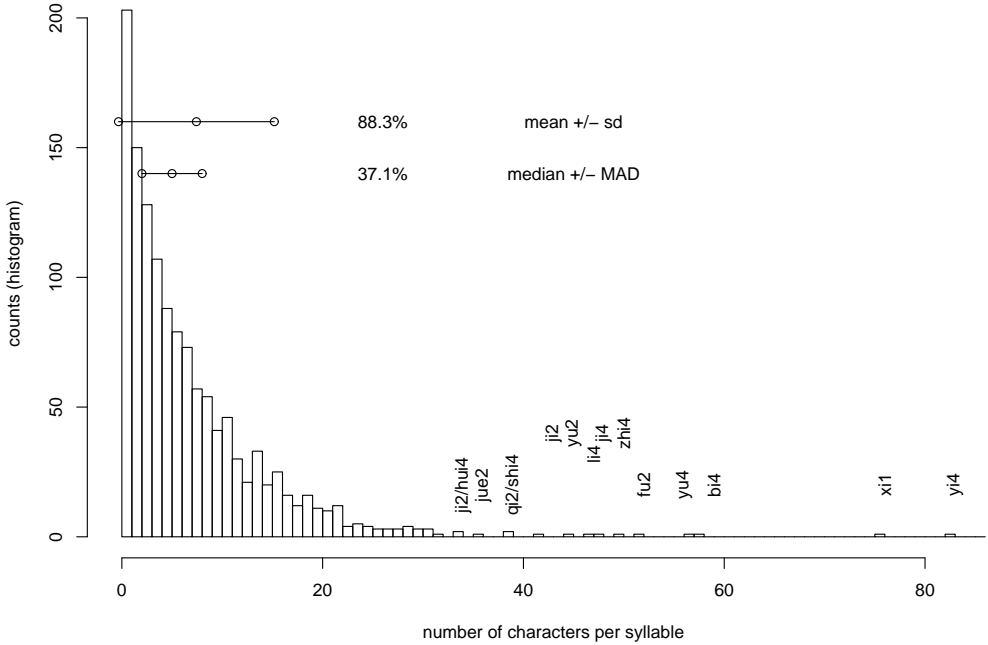, width=8.5cm}
  \end{turn}
\end{center}
\caption{
\label{fig1}
Histogram of $n_c$ (number of characters per syllable). The
$x$-axis ($n_c$ value) are partitioned in bins (bin size is 1),
and $y$-axis is the number of syllables that fall in a bin.
Also marked in the plot are: mean plus/minus one standard deviation
of $n_c$; median plus/minus one MAD of $n_c$; the names of
the top 15 syllables.
}
\end{figure}

\section{Data and Methods}

\indent

{\bf Modern Chinese Small Dictionary:} The 4th edition of
{\sl XianDai HanYu Xiao CiDian} (translated as {\sl Modern
Chinese Small Dictionary}), published by {\sl The Commercial
Press} in 2006, collected 9212 characters, from 1280
intonation-included syllables. The number of toned syllables
is less than $412 \times 4= 1648$ because some intonation
on certain sound do not exist.  No traditional characters are 
included if the corresponding simplified character is already 
in the dictionary. Because some characters can be pronounced in  
multiple ways, the number of uniquely pronounced character count
is 9505.

An intonation-included syllable is written in {\sl pinyin}
followed by one of the four numbers (1,2,3,4) for the four
intonations of the level, raising, dipping, and falling tones. 
For example, hao1, hao2, hao3, hao4, with the consonant  
initial {\sl h}, vowel final {\sl ao}, plus four intonations. 
An alternative notation for the four intonations is 
h\={ao}, h\'{ao}, h\v{ao}, h\`{ao} for the four  tones.

{\bf Logarithmic function and piecewise linear logarithmic
function:} Denote $n_c(r)$ the number of characters per
syllable for rank-$r$ syllable, and rank $r$ ranges from
high ($r=1$)  to low ($r=n$). The $n_c(r)$ can be normalized to
$y_r \equiv n_c(r)/\sum_{r=1}^n n_c(r)$. Given the data
\{ $r$, $y_r$ \} ($r=1, 2, \cdots, n$), a simple fitting
function is the logarithmic function:
\begin{equation}
\label{eq-log}
f_c(r)= C+ a \log(r),
\end{equation}
with one fitting parameter $a$ ($C$ is constrained by the
normalization condition $\sum_{r=1}^n f_c(r)=1$).
A piecewise logarithmic function consists of multiple logarithmic
fitting functions, one for each rank range, e.g.:
\begin{equation} 
f_c(r) =  \left\{
 \begin{array}{ll}
 C + a \log(r) & \mbox{ if $1 \le r \le r_0$ }  \\
 C' + a' \log(r) & \mbox{ if $r_0 < r \le n$ }.  
 \end{array}
\right.
\end{equation} 
where $r_0$ is the partition rank point separating the two
logarithmic functions for high-ranking and low-ranking ranges.
A piecewise function can be continuous or not. To force the
continuity for the above two-piece logarithmic function,
one requires $C+ a\log(r_0)= C'+a'\log(r_0)$. Without
continuity requirement, the number of fitting parameters in the two-piece
logarithmic function is 4: $a, C', a', r_0$, and the
continuity requirement reduces that number by 1, to 3.

{\bf Beta rank function:} The Beta function was proposed in
\citep{beta1,beta2,beta3} for fitting ranked plots:
\begin{equation}
\label{eq-beta}
f_c(r) = C \frac{ (n+1-r)^b}{ r^a},
\end{equation}
where $a$ and $b$ are two fitting parameters (scaling exponents). 
Beta function Eq.(\ref{eq-beta}) is a modification from 
the power-law function where $b=0$.  The best known power-law 
function for ranked data is the Zipf's law of word usages \citep{zipf}. 
Note that Eq.(\ref{eq-beta}) shares a similar form as the Beta 
distribution $p(x)= C x^\alpha (1-x)^\beta$, but it is not a 
probability density distribution.

{\bf Nonlinear regression:} We use the $R$ 
({\sl http://www.r-project.org/}) package {\sl nls}
for non-linear least square procedure used in regressions \citep{nls}.
This non-linear procedure requires an initial value of the
parameters, and the best fit is achieved numerically. 
Choosing the appropriate initial parameter values is important
if there are multiple local optimal solutions. We first
use a linear regression on the transformed variable, $y= \log(C)+a x + b x_2$
(where $y=\log(f), x=-\log(r), x_2= \log(n+1-r)$, 
to obtain an estimation of the parameter values.
Then these estimations of $a$ and $b$ (and $C$) are used as 
the initial condition for the non-linear regression.

{\bf Regression performance:} Given the data $\{ r, y_r \}$
($r=1,2, \cdots n$), the fitting performance of a function 
$f_c(r)$ is measured by sum of squared errors (SSE):
\begin{equation}
SSE \equiv \sum_{r=1}^n (f_c(r) - y_r)^2.
\end{equation}
Although SSE increases with the number of points ($n$),
for the purpose of comparing two or several functions on the
same dataset, the value of $n$ is the same, and it is
not necessary to normalize (divide by $n$) SSE.

{\bf Model selection by AIC:} Since functions with more
fitting parameters should be able to fit the data no worse
than those functions with less number of free parameters,
SSE itself can not be used to compare two functions with
different number of parameters. One method to discount the
effect of extra parameter is to penalize the number of parameters.
The Akaike Information Criterion (AIC) \citep{aic} subtracts
a term from the log-maximum-likelihood that is twice the
number of parameter ($K$). In the regression context, it can be
shown that under the condition of unknown variance of
the noise, AIC is equal to \citep{ripley}:
\begin{equation}
AIC = n \log (SSE/n) + 2K.
\end{equation}
Among different fitting functions, the one with the smallest AIC
is the best model, either due to smaller error SSE or due
to fewer number of parameters $K$.

To compare two AIC's for two different fitting functions,
we have:
\begin{equation} 
\label{eq-del-aic}
AIC_2 - AIC_1 = n \log \frac{SSE_2}{SSE_1} + 2(K_2-K_1).
\end{equation} 
Suppose the second function fits the data better than the
first function but utilizing more parameters, then the first 
term in Eq.(\ref{eq-del-aic}) is negative, but the second term is
positive. Only when the magnitude of the negative term is 
large enough to compensate the second positive term, 
is the second function selected.

{\bf Simulated data from Poisson distribution:}
One can simulate a syllable-to-character mapping spectrum 
based on the observed one. For a syllable with $n_c$ characters,
we can treat $n_c$ as the mean of a Poisson distribution:
Pois($\lambda=n_c$). Note that the standard deviation of 
Poisson distribution is $\sqrt{\lambda}=\sqrt{n_c}$ and the lowest
possible value is 0 (as a comparison, if we use the normal 
distribution $N(\mu=n_c, \sigma=\sqrt{n_c})$ in simulation, 
we may have negative count values).  From the observed 
\{ $n_c(r)$ \} ($r=1,2, \cdots n$), a new set sampled from 
the Poisson distribution \{ $n_c'$ \} is generated.
We then remove points with $n_c'=0$ and re-rank them which
becomes one replicate of the simulated dataset. This process is
repeated 1000 times.

\begin{figure}[th]
\begin{center}
  \begin{turn}{-90}
   \epsfig{file=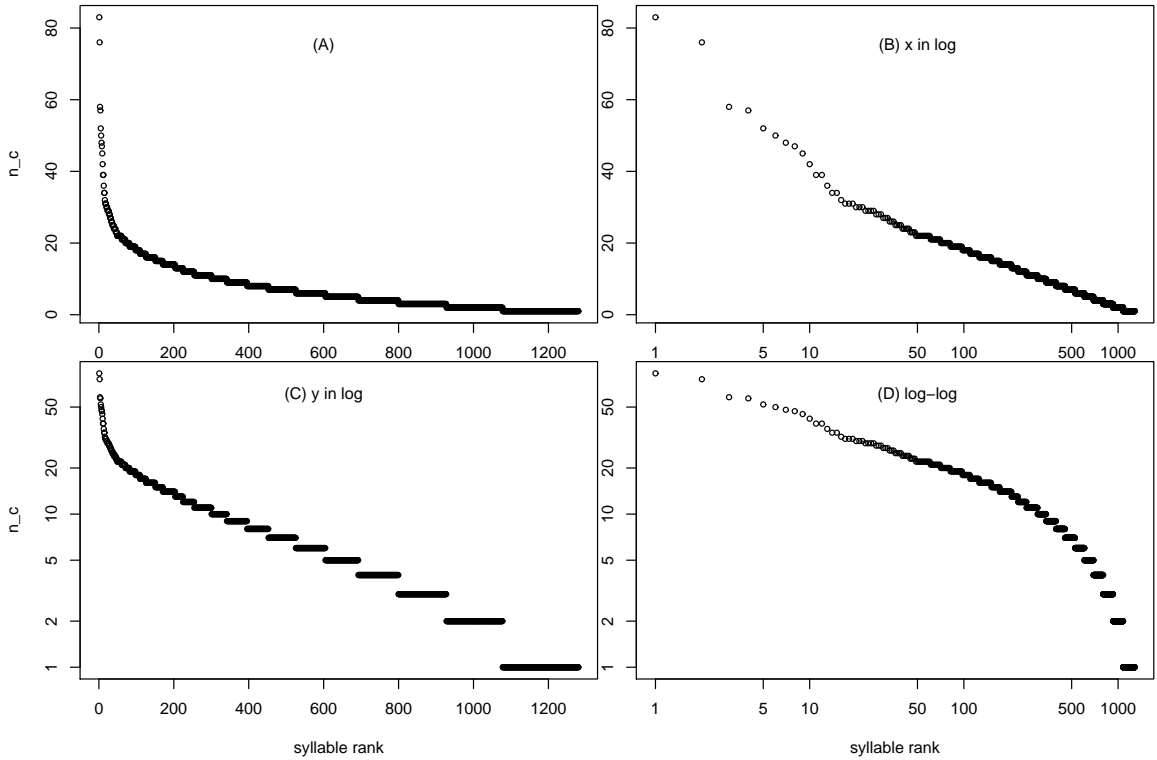, width=10cm}
  \end{turn}
\end{center}
\caption{
\label{fig2}
Ranked syllable-to-character mapping spectrum with $n=$1280
syllables and 9505 characters. The $x$-axis is the rank
of syllable ($r=1$ for most polymorphic syllable, $r=1079$ to 1280
for monomorphic syllables). The same spectrum is displayed
in four different versions: (A) linear-linear; 
(B) ($x$)log-linear; (C) linear-($y$)log; (D) log-log.
}
\end{figure}

\section{Results}

\indent

{\bf Simple Statistics:} The top-ranking syllables with largest
numbers of characters are yi4 (83), xi1 (76), bi4 (58), yu4 (57), fu2 (52).
% The 83 characters pronounced as yi4 are listed in Table \ref{table1}.
The next 10 polymorphic syllables (rank $r=6$ to 15) are
zhi4 (50), ji4 (48), li4 (47), yu2 (45), ji1 (43), qi2 (39), shi4 (39),
jue2 (36), ji2 (34), hui4 (34).  There are 203 syllables with 
only one character (one-to-one, monomorphic). The mean number 
of characters per toned syllable ($n_c$) is 7.4 (=9505/1280), 
median number is 5, standard deviation is 7.6, and 
median-absolute-deviation (MAD) is 3. The range mean $\pm$
standard deviation (0, 15.185) covers 88.3\% of all syllables,
whereas the range median $\pm$ MAD (2, 8) covers only 37.1\%.

Fig.\ref{fig1} shows the histogram of the number of characters per
syllable $n_c$, which summarizes all the above statements in a graph. 
Note that the statistical numbers will be dictionary-dependent
as governed by the ``large number of rare event" model \citep{baayen},
but the shape of the histogram in Fig.\ref{fig1} is
expected to remain similar.

Another way to describe the uneven distribution of
the number of characters per syllable $n_c$ is  how much the characters
are covered under the top $x$\% of the most polymorphic
syllables. For example, top 1\% of top syllables contain 7\%
of all characters, 5\% contain 21\%, 10\% contains 33\%, 
and top 25\% contains 59\%.  The ``few polymorphic syllables but
large amount of monomorphic syllables", non-Gaussian-like \citep{siam}, 
distribution in Fig.\ref{fig1} also points to an uneven rank distribution.

{\bf Ranked distribution of the number of characters per syllable:}
Like so many phenomenological laws in quantitative linguistics,
a better and smoother description of the data is by ranking the data points
in order \citep{wli-entropy}. We rank syllables by their $n_c$ 
values, from large to small, and the resulting rank plot is in 
Fig.\ref{fig2} in four different versions: regular (linear-linear), 
($x$)log-linear, linear-($y$)log, and log-log. 
Fig.\ref{fig2}(A) and Fig.\ref{fig2}(C) shows
that the rank function is neither linear nor exponential.
Fig.\ref{fig2}(B) indicates that a logarithmic or piece-wise
logarithmic function might be good fitting functions.
Fig.\ref{fig2}(D) hints that a power-law function can not
fit the data well, but a modified one, such as the Beta rank
function, might fit the data better \citep{wli-physicaa}.

Related to the uneven distribution of $n_c$, Fig.\ref{fig2} 
can also be converted to a Lorenz curve (not shown) for 
calculating the Gini coefficient. In Lorenz curve,
the $x$-axis is a cumulation of the number of syllables (from
0 to 100\%) and $y$-axis is a cumulation of the number of
characters (also from 0 to 100\%). If all syllables have  
the same number of characters, the Lorentz curve will overlap
with the diagonal line. Gini coefficient ($G$) is defined as twice
the area between the diagonal line and the Lorenz curve.
$G=0$ means a complete equality among syllables, and $G=1$
the complete inequality. The Gini coefficient for this dataset
is $G=0.49$ by the function {\sl gini} in the {\sl R} {\sl reldist}
package \citep{reldist}. This function is an implementation of
the formula: $G=n^{-1} (n+1 - 2(\sum_r (n+1-r) n_c(r)/\sum_r n_c(r) )) $.

\begin{figure}[th]
\begin{center}
  \begin{turn}{-90}
   \epsfig{file=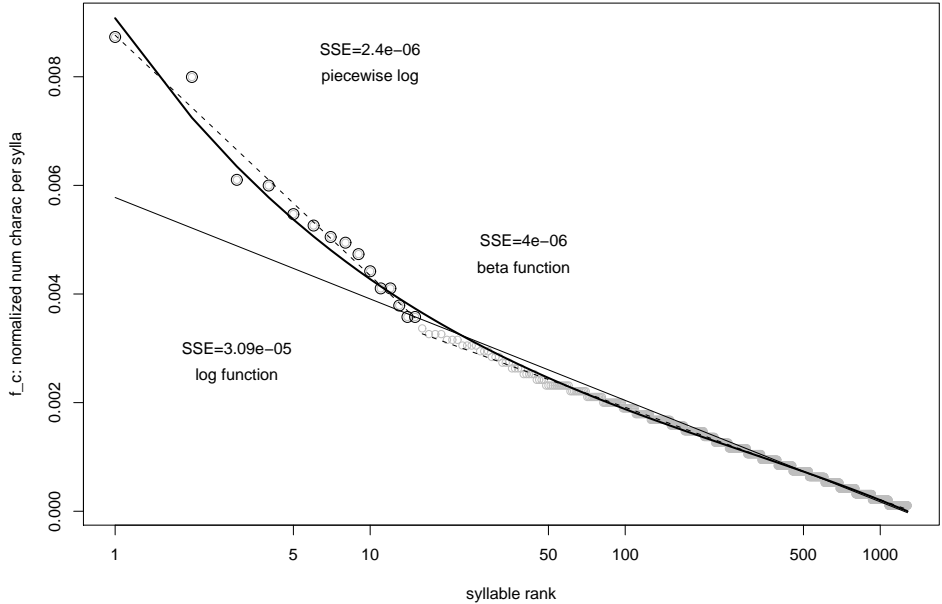, width=8cm}
  \end{turn}
\end{center}
\caption{
\label{fig3}
Ranked syllable-to-character mapping spectrum with $x$ (rank)
in logarithmic scale. Three fitting functions are superimposed
to the data: (1) logarithmic function (thin solid line);
(2) Beta rank function (thick solid line);
(3) piecewise logarithmic function with the partition
point $r_0=15$.
}
\end{figure}

{\bf Fitting the ranked distribution of number of characters per syllable --
logarithmic function and piecewise logarithmic function:} 
The best fit logarithmic function of ranked
$y_r= n_c(r)/\sum_r n_c(r)$ shown in Fig.\ref{fig3} is
$f_c(r) =5.78 \times 10^{-3} - 8.11 \times 10^{-4} \log(r) $.
From Fig.\ref{fig3}, the fitting function clearly under-estimates
the $f_c$ for top-ranking syllables, which is compensated by
over-estimation of $f_c$ at low-ranking tails (there are many
more low-ranking syllables than the high-ranking ones).

Fig.\ref{fig3} strongly suggests that the data can be split into 
two rank ranges, one for high-ranking syllables (from $r=1$ to $r=15$)
and another for the rest ($r > 15$).  Interestingly, the
histogram of $n_c$ in Fig.\ref{fig1} does show that top 15
syllables (yi4, xi1, bi4, yu4, fu2, zhi4, ji4, li4, yu2, 
ji1, qi2, shi4, jue2, ji2, bui4) seem to be outliers with 
respect to the more continuous distribution for other 
syllables. For simplicity, we apply the two-piecewise 
logarithmic functions without requiring continuity. 
The logarithmic function fitting the top 15 syllables is 
$f_c(r)= 0.00877- 0.00192 \log(r)$, and that for the rest 
is $f_c(r)= 0.00532 - 0.000739 \log(r)$.  Indeed, the slope 
of the first function is steeper than the second.

{\bf Fitting the ranked distribution of number of characters per syllable --
Beta rank function:} 
The less than perfect fitting of the logarithmic
function in Fig.\ref{fig3} calls for the use of two-parameter
functions, as previously advocated by us in \citep{wli-entropy}.
The Beta function \citep{beta1,beta2,beta3} has been proven
to be a robust function that fits well diverse types of
ranked linguistic data \cite{wli-entropy}. Fig.\ref{fig3} shows
the best-fit Beta function by non-linear regression (in thick solid line):
\begin{equation}
f_c(r) = 5.95 \times 10^{-6} \frac{ (1281- r)^{1.025}}{r^{0.324}}.
\end{equation}
The fact that the fitting value $b=1.025$ is larger than $a=0.324$
indicates that unlike the situation of Zipf's law, the power-law 
function is not a good fitting function of this data.
More discussion on the meaning of relative magnitude of $a$ and
$b$ can be found in \citep{roberto}, and discussion on the
range-limited rank variable versus range-open rank variable
can be found in \citep{wli-physicaa}.

{\bf Comparison of curve fitting performance and model
selection:} Both Beta function and piecewise logarithmic function 
seem to  fit the data better than the one-piece logarithmic function. 
To quantify the fitting performance, we use SSE as the measure
of discrepancy and AIC to compare models. SSE's of piecewise
logarithmic, Beta, and logarithmic functions are
$2.36 \times 10^{-6}$, $3.95 \times 10^{-6}$ and $3.09 \times 10^{-5}$
respectively, confirming that piecewise logarithmic and Beta
function are both better than the logarithmic function, with
piecewise logarithmic function even better.

The low SSE value for the piecewise logarithmic function is
not affected by the continuity requirement. If the continuity
condition is imposed and the low-ranking syllables are fitted
first, SSE can be 2.76 $\times 10^{-6}$, 2.63 $\times 10^{-6}$,
or 2.53 $\times 10^{-6}$ if the converging point is at $r=15,
15.5, 16$. If the high-ranking syllables are fitted first,
then SSE values are much worse (6.80 $\times 10^{-6}$, 5.56 $\times 10^{-6}$,
4.54 $\times 10^{-6}$).

The calculation of AIC is tricky for the piecewise logarithmic
function. Even though $r_0=15$ is not fitted by the data,
it is however chosen by inspecting Fig.\ref{fig2}(B). If
we use the AIC of Beta function ($K=3$) as the baseline value,
AIC of logarithmic function ($K=2$) is larger by 2629,
and AIC of piecewise logarithmic function (without continuity)
($K=4$) is smaller by 659. The first term in Eq.(\ref{eq-del-aic}),
$1280 \times \log(2.3554/3.9534)= -662$ is so large that
even if the more severe penalty of model complexity is
imposed (e.g. Bayesian information criterion, BIC \citep{bic}), 
the piecewise logarithmic function is still selected.

\begin{figure}[th]
\begin{center}
  \begin{turn}{-90}
   \epsfig{file=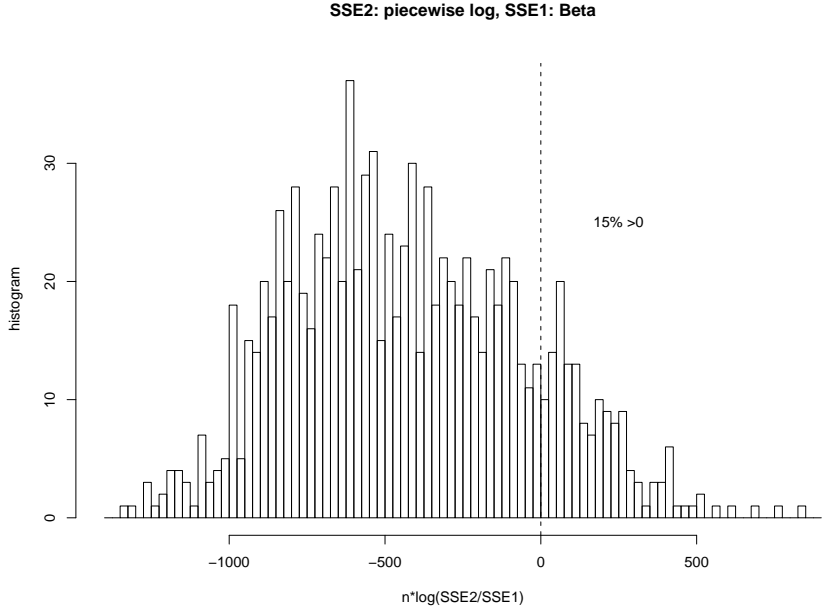, width=8cm}
  \end{turn}
\end{center}
\caption{
\label{fig4}
Histogram of $n \cdot \log(SSE2/SSE1)$ of 1000 replicates simulated
by the observed-data-based Poisson distribution, where $n$ is
the number of syllables with nonzero count of number of characters,
SSE2 is the sum of squared errors for the two-piece logarithmic function,
and SSE1  is that for the Beta rank function.
}
\end{figure}

{\bf Noise and significance:} To check the robustness of
the model comparison result, we create replicates of simulated
data that are closely related to the observed one, in order to
check how often the same result holds. The proportion of replicates
that lead to the opposite conclusion from the observed data is
the empirical $p$-value. The simulated data is sampled from the
Poisson distribution with the observed number-of-character per
syllable ($n_c$) as the mean. It is well known that the mean
and the variance of the Poisson distribution is the same, thus
the standard deviation is $\sqrt{n_c}$.

For each replicate of Poisson distribution based
syllable-to-character mapping spectrum, SSE from the Beta rank
function and SSE from the piecewise logarithmic function are
calculated. For the SSE from the Beta rank function, we again
use the nonlinear regression with the results from linear regression
as the initial condition. For SSE from the piecewise logarithmic
functions, we treat the segmentation rank $r_0$ as a running variable,
from 2 to 1/5 of the maximum rank, and the $r_0$ with the best SSE 
is chosen.  The continuity between the two logarithmic functions is 
enforced, and the high-ranking syllables are fitted first.

Fig.\ref{fig4} shows the histogram, from the simulation by
Poisson distribution, of the first term in the AIC difference 
(Eq.(\ref{eq-del-aic})) between the two fitting functions:
$n \cdot \log (SSE2/SSE1)$, where SSE2 is for the two-piece logarithmic 
function, SSE1 for the Beta rank function, and $n$ the number of
syllables with nonzero number of characters (due to random
sampling, a Pois($\lambda=1$) distribution has a high chance
to sample a zero value).  The majority of
these values are negative, and 15\% are positive. Since $K_2-K_1$
is only 1, it can be shown from Fig.\ref{fig4} that 
at most 1\% of the replicates have any chance to
compensate the negative $n \cdot \log (SSE2/SSE1)$ term
by larger 2$ \dot (K_2-K_1)$ for AIC or larger $\log(n) \cdot (K_2-K_1) $ 
for BIC. This leaves 0.16 as our rough estimation of the 
$p$-value, or the significance in testing the hypothesis that
two-piece logarithmic function is better than the Beta rank function.

\section*{Discussions}

\indent

Both spoken and written languages change with time. It
is pointed out in \citep{wang73} that spoken Chinese evolves
with a faster speed than written Chinese. Thus a character 
may be pronounced in a different way in ancient Chinese 
from that in modern Chinese. The drifting of pronunciation may 
introduce a flux, both in and out, in $n_c$ values. 
On the other hand, ancient characters are often out 
of favor in modern Chinese. This however may not cause a 
problem because the rarely used characters are still in the dictionary.
How syllable-to-character mapping spectrum change with time
is an interesting question of which we do not yet know an answer.

The conclusion reached in \citep{wli-physicaa}, that
the Beta rank function fits the syllable-to-character mapping spectrum
better than the logarithmic function remains true when
intonation of syllables is considered, despite a 4-fold
expansion in the $x$-axis. However, we have a new conclusion
that the piecewise logarithmic function fits the data even
better than the Beta rank function. 

Although this conclusion seems to be solid by a rigorous 
model selection technique, the difference of fitting 
performance between the two is actually very small 
(2.36 $\times 10^{-6}$ vs. 3.95 $\times 10^{-6}$).
The rank range ($r=1$ to $r=15$) covered by the second
logarithmic function is only 1.2\% of the total in
linear scale and 37.8\% in log-scale. These can be used
in an argument against the claim that the outperformance
by the piecewise logarithmic function is real.
Indeed, the empirical $p$-value of 0.15-0.16
obtained by simulation via a Poisson distribution, shows
that the statistical evidence is relatively weak, as
the standard criterion for rejecting a hypothesis, with
the implication of accepting another hypothesis, is
to have a $p$-value smaller than 0.5.

The relative fitting performance can be sensitive to small 
changes in the data.  We notice that the high-ranking 
syllables are outliers in the distribution of $n_c$ in
Fig.\ref{fig1}. Outliers tend to be not reproducible,
or have larger variability, which implies we may not
reproduce the same 15 $n_c$ values in a replicated run.
The main intention of our simulation is
to randomize the value of outliers by a Posson distribution.
Examining the top-ranking syllable yi4 in Table 1, for
example, shows that only 24 or so characters are more
commonly used in the modern Chinese, others are mostly
obscure characters. Some are so rare that a Chinese reader may 
not encounter them in her lifetime. Using an even smaller
dictionary will reduce $n_c$ for all syllables, of
course, but the effect on the top ranking $n_c$'s is
less predictable.

In conclusion, we would like to show that it is interesting
to study the functional form of the syllable-to-character
spectrum which connects the spoken and written Chinese.
While two-piece logarithmic functions apparently outperforms
the Beta rank function in our data, simulation shows that
15\%-16\% of the time the Beta function may outperform
the two-piece logarithmic function.

\section*{Acknowledgements}

I would like to thank Jianmin Bai, Pedro Miramontes for helpful comments
and suggestions.

\end{document}